\pdfoutput=1

\documentclass[11pt]{article}

\usepackage[final]{emnlp2021}

\usepackage{times}
\usepackage{latexsym}

\usepackage[T1]{fontenc}

\usepackage[utf8]{inputenc}

\usepackage{microtype}

\usepackage{booktabs}
\usepackage{makecell}
\usepackage{graphicx}
\usepackage{amssymb}
\usepackage{url}
\usepackage{float}

\usepackage{header}
\usepackage[font={small}]{caption}

\newcommand{\ourtitle}{Systematic Generalization on gSCAN: \\
What is Nearly Solved and What is Next?}
\usepackage{amsmath,amsfonts,amssymb}
\usepackage{xspace}

\providecommand{\eg}{\textit{e.g.}\@\xspace}
\providecommand{\ie}{\textit{i.e.}\@\xspace}

\newcommand{\vh}{\mathbf{h}}

\newcommand{\vx}{\mathbf{x}}

\newcommand{\mH}{\mathbf{H}}

\newcommand{\mO}{\mathbf{O}}

\newcommand{\mR}{\mathbf{R}}

\newcommand{\mT}{\mathbf{T}}

\newcommand{\mV}{\mathbf{V}}
\newcommand{\mW}{\mathbf{W}}

\newcommand{\Rbb}{\mathbb{R}}

\newcommand{\RN}[1]{\uppercase\expandafter{\romannumeral#1}}

\newcommand{\footremember}[2]{%
   \thanks{\xspace\xspace#2}
    \newcounter{#1}
    \setcounter{#1}{\value{footnote}}%
}
\title{\ourtitle}

\author{
  Linlu Qiu$^1$\footremember{aires}{Work done as a Google AI Resident.} 
  ~~~~Hexiang Hu$^1$\footremember{Google}{Work done while at USC.} 
  ~~~~Bowen Zhang$^2$ 
  ~~~~Peter Shaw$^1$ 
  ~~~~Fei Sha$^1$ \\
    \\
  $^1$Google Research~~~~~~~~~~$^2$USC
}

\begin{document}
\maketitle

\begin{abstract}
We analyze the \emph{grounded} SCAN (gSCAN) benchmark, which was recently proposed to study systematic generalization for grounded language understanding. First, we study which aspects of the original benchmark can be solved by commonly used methods in multi-modal research. We find that a general-purpose Transformer-based model with cross-modal attention achieves strong performance on a majority of the gSCAN splits, surprisingly outperforming more specialized approaches from prior work. Furthermore, our analysis suggests that many of the remaining errors reveal the same fundamental challenge in systematic generalization of linguistic constructs regardless of visual context. Second, inspired by this finding, we propose challenging new tasks for gSCAN by generating data to incorporate relations between objects in the visual environment. Finally, we find that current models are surprisingly data inefficient given the narrow scope of commands in gSCAN, suggesting another challenge for future work.

\end{abstract}

\section{Introduction}
Systematic generalization refers to the ability to understand new compositions of previously observed concepts and linguistic constructs. While humans  exhibit this ability, neural networks often struggle.
To study systematic generalization, several synthetic datasets have been proposed. \citet{lake2018generalization} introduced SCAN, a dataset composed of natural language instructions paired with action sequences, and splits in various ways to assess systematic generalization. Recently, ~\citet{ruis2020benchmark} introduced the grounded SCAN (gSCAN) benchmark, which similarly pairs natural language instructions with action sequences, but further requires that instructions are interpreted within the context of a grid-based visual navigation environment. In this work, we analyze which aspects of gSCAN can currently be solved by general-purpose models, and propose new tasks and evaluation metrics for future research of systematic generalization on gSCAN.

First, to understand which aspects of gSCAN can be addressed by a general-purpose approach, we evaluate a Transformer-based model with cross-modal attention. Cross-modal attention has been proven effective for other multi-modal tasks~\cite{lu2019vilbert,tan2019lxmert,chen2020uniter}. 
It achieves strong performance on a majority of the splits, surprisingly outperforming several ``specialist'' approaches designed for gSCAN~\citep{heinze2020think,gao2020systematic, kuo2020compositional}.
We analyze the remaining errors, finding that many of the errors appear to be related to the same fundamental challenge in systematic generalization of linguistic constructs studied in datasets such as SCAN, regardless of the visual context.

Our analysis motivates the creation of an additional gSCAN task, which features a greater degree of complexity in how natural language instructions are grounded in the visual context.
In this task,  the agent needs to reason about spatial relations between objects expressed in language. We find this new task to be challenging for existing models.

Finally, we also assess the data efficiency of our cross-modal attention model on gSCAN.
We find that despite the simplicity of the world state and the grammar used to generate instructions, model performance on most splits declines significantly when provided with less than $\sim$40\% of the 360,000 original training examples. This suggests that we should consider sample complexity for future work.

\section{Cross-modal Attention Solves gSCAN, Almost}

\begin{table*}
\centering
{
\small
\tabcolsep 3pt
\begin{tabular}{@{\;}ccccccccc@{\;}}
    \toprule
    Split & Seq2Seq~\shortcite{ruis2020benchmark} & GECA~\shortcite{andreas2020good} & Kuo~\shortcite{kuo2020compositional} & Heinze~\shortcite{heinze2020think} & Gao~\shortcite{gao2020systematic} & FiLM~\shortcite{perez2018film} & RN~\shortcite{santoro2017simple} & Ours \\
    \midrule
    A & 97.69 \scriptsize{{$\pm$ 0.22}} & 87.6 \scriptsize{$\pm$ 1.19} & 96.73 \scriptsize{$\pm$ 0.58} & 94.19 \scriptsize{$\pm$ 0.71} & 98.60 \scriptsize{$\pm$ 0.95}  & 98.83 \scriptsize{$\pm$ 0.32} & 97.38 \scriptsize{$\pm$ 0.33} & \textbf{99.95 \scriptsize{$\pm$ 0.02}} \\
    B & 54.96 \scriptsize{$\pm$ 39.3}9 & 34.92 \scriptsize{$\pm$ 39.3}0 & 94.91 \scriptsize{$\pm$ 1.30} & 87.31 \scriptsize{$\pm$ 4.38} & 99.08 \scriptsize{$\pm$ 0.69} & 94.04 \scriptsize{$\pm$ 7.41} & 49.44 \scriptsize{$\pm$ 8.19} & \textbf{99.90 \scriptsize{$\pm$ 0.06}} \\
    C & 23.51 \scriptsize{$\pm$ 21.8}2 & 78.77 \scriptsize{$\pm$ 6.63} & 67.72 \scriptsize{$\pm$ 10.8}3 & 81.07 \scriptsize{$\pm$ 10.1}2 & 80.31 \scriptsize{$\pm$ 24.5}1 & 60.12 \scriptsize{$\pm$ 8.81} & 19.92 \scriptsize{$\pm$ 9.84} & \textbf{99.25 \scriptsize{$\pm$ 0.91}} \\
    D & 0.00 \scriptsize{$\pm$ 0.00} & 0.00 \scriptsize{$\pm$ 0.00} & \textbf{11.52 \scriptsize{$\pm$ 8.18}} & - & 0.16 \scriptsize{$\pm$ 0.12} & 0.00 \scriptsize{$\pm$ 0.00} & 0.00 \scriptsize{$\pm$ 0.00} & 0.00 \scriptsize{$\pm$ 0.00} \\
    E & 35.02 \scriptsize{$\pm$ 2.35} & 33.19 \scriptsize{$\pm$ 3.69} & 76.83 \scriptsize{$\pm$ 2.32} & 52.8 \scriptsize{$\pm$ 9.96} & 87.32 \scriptsize{$\pm$ 27.3}8 & 31.64 \scriptsize{$\pm$ 1.04} & 42.17 \scriptsize{$\pm$ 6.22} & \textbf{99.02 \scriptsize{$\pm$ 1.16}} \\
    F & 92.52 \scriptsize{$\pm$ 6.75} & 85.99 \scriptsize{$\pm$ 0.85} & 98.67 \scriptsize{$\pm$ 0.05} & - & 99.33 \scriptsize{$\pm$ 0.46} & 86.45 \scriptsize{$\pm$ 6.67} & 96.59 \scriptsize{$\pm$ 0.94} & \textbf{99.98 \scriptsize{$\pm$ 0.01}} \\
    G & 0.00 \scriptsize{$\pm$ 0.00} & - & \textbf{1.14 \scriptsize{$\pm$ 0.30}} & - & - & 0.04 \scriptsize{$\pm$ 0.03} & 0.00 \scriptsize{$\pm$ 0.00} & 0.00 \scriptsize{$\pm$ 0.00} \\
    H & 22.70 \scriptsize{$\pm$ 4.59} & 11.83 \scriptsize{$\pm$ 0.31} & 20.98 \scriptsize{$\pm$ 1.38} & - & \textbf{33.6 \scriptsize{$\pm$ 20.8}1} & 11.71 \scriptsize{$\pm$ 2.34} & 18.26 \scriptsize{$\pm$ 1.24} & 22.16 \scriptsize{$\pm$ 0.01} \\
    \bottomrule
\end{tabular}
}
\caption{Percentage of exact matches by various methods on gSCAN's compositional splits.}
\label{tab:exact_match}
\end{table*}
\begin{table}
\centering
\scriptsize
\tabcolsep 4pt
{
    \begin{tabular}{p{0.3\linewidth} p{0.6\linewidth}}
    \toprule
    Split & Held-out Examples \\
    \midrule
    A: Random & Random \\
    B: Yellow squares & \emph{Yellow squares} referred with color and shape \\
    C: Red squares & \emph{Red squares} as target \\
    D: Novel direction & Targets are located \emph{south-west} of the agent \\
    E: Relativity & Circles with size 2 and referred as \emph{small} \\
    F: Class inference & Squares with size 3 and need to be pushed twice \\
    G: Adverb k = 1 & Commands with \emph{cautiously} \\
    H: Adverb to verb & Commands with \emph{while spinning} and \emph{pull} \\
    \bottomrule
    \end{tabular}
}
\caption{Held-out examples on gSCAN's compositional splits.}
\label{tab:split_details}
\end{table}

\paragraph{Experimental Setup} gSCAN has two types of generalization tasks: compositional generalization (CG) and length generalization.
We focus on CG splits, which consist of a shared training set, 1 random test set, and 7 test sets of specifically designed held-out examples such that examples differ from the training set in various ways, cf. Table~\ref{tab:split_details}.
The input commands (\eg \emph{walk to a red circle hesitantly}) are synthetically generated. The agent observes the world state which is \fs{perhaps directly state how big the grid space is?} \llq{$d$ can be any number. The original one uses a larger grid space for target lengths split. I added more info as we only use 6}a $d \times d$ grid ($d=6$ in our case) with objects in various visual attributes. The output is an action sequence in the grid world (\eg \emph{turn left, walk, walk, stay}).
We use exact match of the entire sequence as evaluation metric and report the mean and standard deviation across 5 runs.

\paragraph{Our Model} We implement a seq2seq model with 6 Transformer layers for the encoder and the decoder each. The architecture is similar to ViLBERT~\citep{lu2019vilbert}, a popular multi-modal model for visual and text information fusion. On a high-level, the encoder's text stream reads the commands while its visual stream encodes the world states. There are \textbf{cross-modal attention} between the two streams. The details are in Appendix~\ref{sec:model}.

\paragraph{Results}
Table~\ref{tab:exact_match} shows the results by various models. Among them,  FiLM and Relation Networks (RN) have achieved strong performance on other synthetic datasets for similar multi-modal tasks~\citep{perez2018film,santoro2017simple}. We also compare to models with specialized designs for CG or even gSCAN: a seq2seq model using data augmentation~\citep{andreas2020good}, auxiliary loss~\citep{heinze2020think}, task-specific architecture~\citep{kuo2020compositional,gao2020systematic}.

\llq{Added discussions from rebuttal for camera ready}The cross-modal attention model outperforms others on 5 out 8 splits. We hypothesize it is more effective as we leverage cross-modal attention to allow bi-directional interaction between language instruction and visual environment, in contrast to prior works that only have uni-directional attention from text to visual context~\citep{ruis2020benchmark, kuo2020compositional, gao2020systematic, heinze2020think}.
The additional attention from visual context to text improves grounding natural language instructions to the visual environment. However, on the ``hard'' splits (D, G and H), all methods struggle.

\begin{figure}[t]
\centering
\includegraphics[width=0.98\linewidth]{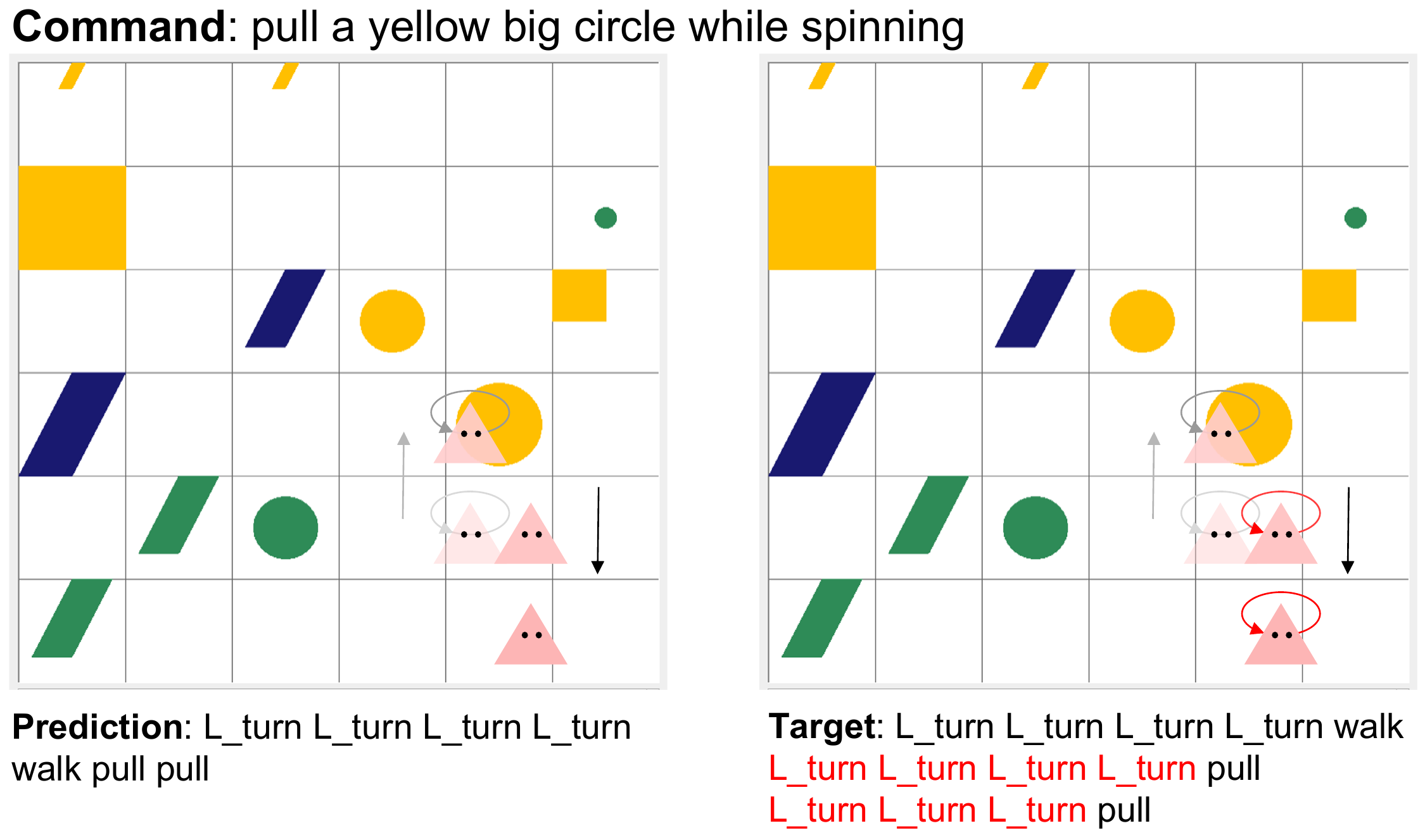}
\caption{A failure case from the H split (Adverb to verb).  Prediction (left) differs from target (right) in action sequences, while correct object is identified. Missing actions are in red.}
\label{fig:failure_case}
\end{figure}

\begin{figure}[t]
    \centering
    \includegraphics[width=0.98\linewidth]{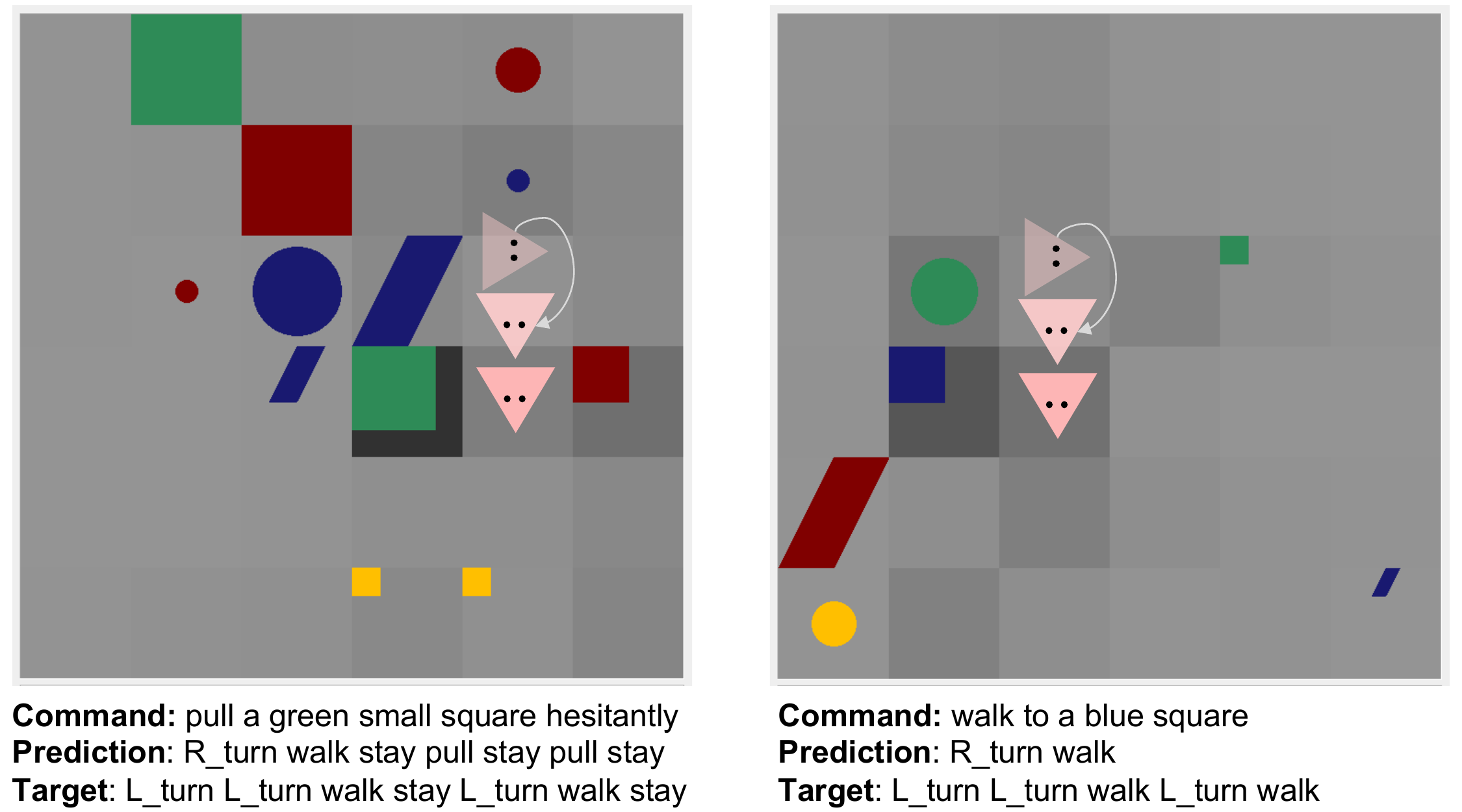}
    \caption{Visualizations of predictions with attention weights for novel direction (D) split. The darker the grid, the higher the attention score.}
    \label{fig:attention_visualization}
\end{figure}

\paragraph{Analysis}
\label{subsec:error_analysis}
First, we analyze in details the ``hard'' splits. Certain aspects of interpreting instructions are highly dependent on visual grounding. Most prominently, every instruction requires resolving the location of a referred object. However, adverbs such as \emph{cautiously} or \emph{while spinning} have the same meaning regardless of the visual context. Figure~\ref{fig:failure_case} shows one such example from the H split. The agent can successfully locate the target object but fails to combine the seen verb \emph{pull} and adverb \emph{while spinning} to generate the correct action sequence. Therefore, to assess the degree to which errors are caused by incorrect visual grounding, we calculate the exact match between the target position and the agent's final position. The results are shown in Table~\ref{tab:error_analysis}. Despite the low exact match of entire sequences, the agent can find the correct targets more than 90\% of time for two adverb splits (G, H). For the novel direction split (D), the direction "south west" is not seen during training. Similarly to ~\citet{ruis2020benchmark}, we find that the correct row \emph{or} column is often selected, but not both. By analyzing attention weights, ~\citet{ruis2020benchmark} attribute this as a failure to generate novel combinations of actions, not necessarily to identify the correct target object location, similar to our findings from visualizing attention weights in Figure~\ref{fig:attention_visualization}.\llq{Moved figure from appendix} Therefore, we hypothesize that for these three splits (D, G, and H), the primary remaining challenge for the cross-modal attention model is not necessarily related to visual grounding.

\begin{table}
\centering
\small
\tabcolsep 2.5pt
{
    \begin{tabular}{@{\;}c@{\;\;}cccc@{\;}}
    \toprule
    Split & Col Match$\uparrow$  & Row Match$\uparrow$ & Full Match$\uparrow$ & No Match$\downarrow$ \\
    \midrule
    D & 32.91 \scriptsize{$\pm$ 3.02} & 45.30 \scriptsize{$\pm$ 6.46} & 0.00 \scriptsize{$\pm$ 0.00} & 21.78 \scriptsize{$\pm$ 3.58} \\
    G & 98.43 \scriptsize{$\pm$ 3.40} & 98.82 \scriptsize{$\pm$ 2.52} & 97.91 \scriptsize{$\pm$ 4.50} & 0.66 \scriptsize{$\pm$ 1.42} \\
    H & 91.23 \scriptsize{$\pm$ 8.53} & 92.90 \scriptsize{$\pm$ 6.54} & 88.16 \scriptsize{$\pm$ 11.40} & 4.04 \scriptsize{$\pm$ 3.71} \\
    \bottomrule
    \end{tabular}
}
\caption{Percentage of exact matches of target position on splits the cross-modal attention model fails. For the novel direction split (D), the agent can end up at correct row or column around 80\% of the time. For the adverb splits (G, H), the agent can usually find the object, but fails to generate action sequence with correct manner.}
\label{tab:error_analysis}
\end{table}

\begin{table}[t]
\centering
{
\small
\tabcolsep 2pt
\begin{tabular}{@{\;}ccccc@{\;}}
\toprule
Split &  Ours & $-$V & $-$T & $-$X \\
\midrule
A  & 99.95 \scriptsize{$\pm$ 0.02} & 99.73 \scriptsize{$\pm$ 0.27} & \textbf{99.96 \scriptsize{$\pm$ 0.01}} & 96.69 \scriptsize{$\pm$ 4.32} \\
B & \textbf{99.90 \scriptsize{$\pm$ 0.06}} & 97.55 \scriptsize{$\pm$ 2.57} & 95.65 \scriptsize{$\pm$ 5.48} & 31.42 \scriptsize{$\pm$ 11.39} \\
C & \textbf{99.25 \scriptsize{$\pm$ 0.91}} & 97.71 \scriptsize{$\pm$ 1.69} & 96.07 \scriptsize{$\pm$ 4.61} & 32.39 \scriptsize{$\pm$ 6.95} \\
E  & \textbf{99.02 \scriptsize{$\pm$ 1.16}} & 87.67 \scriptsize{$\pm$ 10.69} & 96.04 \scriptsize{$\pm$ 5.15} & 43.19 \scriptsize{$\pm$ 22.82} \\
F & \textbf{99.98 \scriptsize{$\pm$ 0.01}} & 99.85 \scriptsize{$\pm$ 0.24} & \textbf{99.98\scriptsize{ $\pm$ 0.01}} & 96.50 \scriptsize{$\pm$ 4.72} \\
\bottomrule
\end{tabular}
}
\caption{Ablation studies of our model on gSCAN (-V, -T, and -X: removing decoder's visual, decoder's textual, and encoder's cross-modal attentions respectively).}
\label{tab:gscan_attention}
\end{table}
Next, we analyze the splits (A, B, C, E, and F) where the cross-modal attention model performs well. Table~\ref{tab:gscan_attention} contrasts several variants of our model. We notice that removing decoder's textual attention (\ie, $-$T) has virtual no significant change. On the other end, removing visual attention ($-$V) causes significantly worsened performance on E split. Removing cross-modal attention ($-$X) significantly degrades all splits, highlighting the benefit of cross-modal attention.

\section{Grounded Spatial Relation CG}

\begin{figure}[t]
\centering
\includegraphics[width=0.95\linewidth]{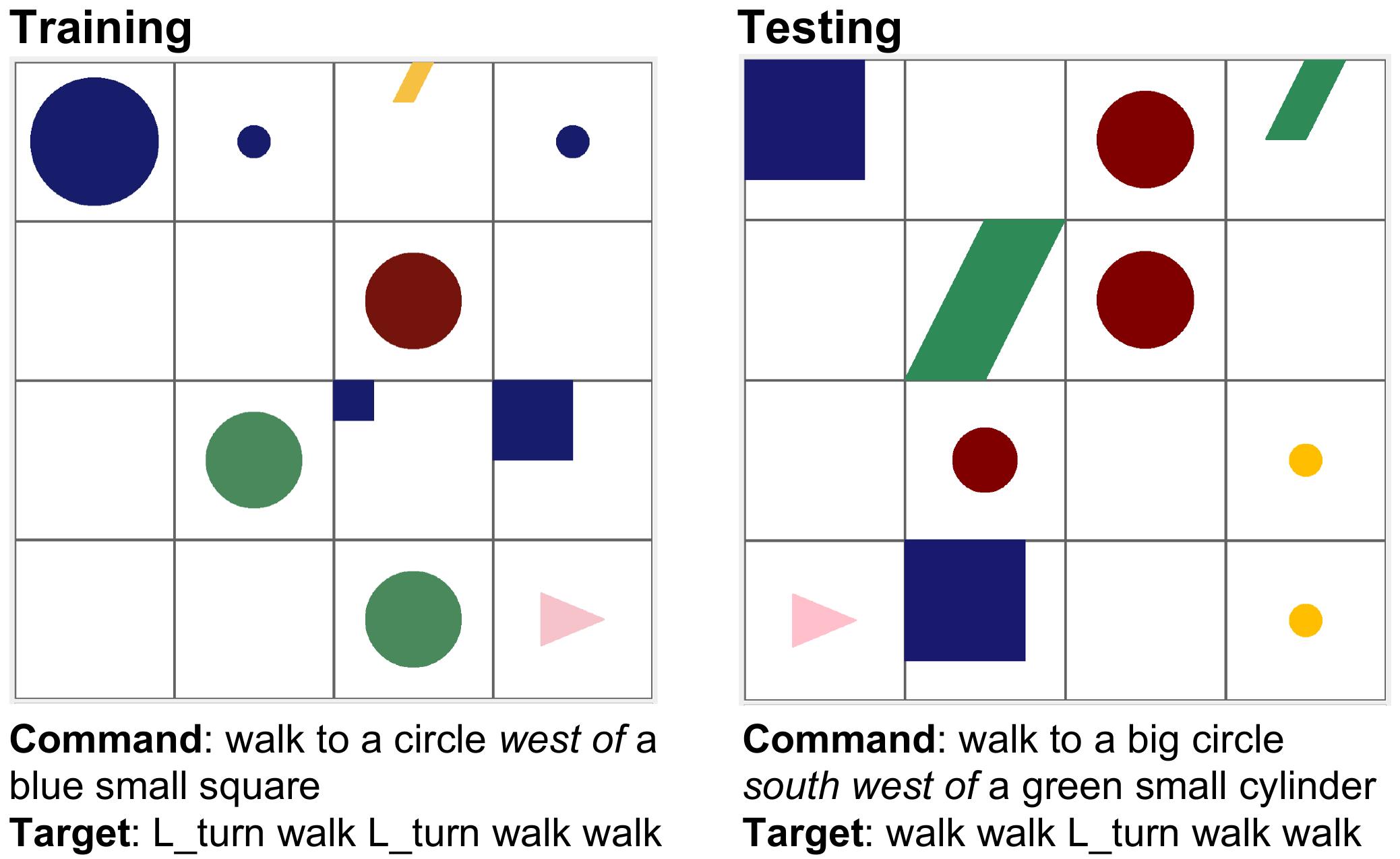}
\caption{\small Training and testing examples for spatial relation splits. The agent is shown as a pink triangle. The agent has seen relative position ``west" (left) and ``south", ``east'', ``south east'' (in other examples) during training and needs to generalize to ``south west" (right) during testing.}
\label{fig:split}
\end{figure}

\paragraph{Proposed Task} Given our analysis suggesting that many of the remaining challenges for gSCAN may not necessarily be related to visual grounding, we propose an additional task that features a greater degree of complexity in how natural language instructions are grounded in the visual environment. We hope this will complement the original gSCAN tasks as a useful assessment of systematic generalization in grounded language understanding.

The new data we create will contain language expressions that refer to target objects along with their relations to a second referenced object. We use two types of relations: \emph{next to} and relative positions such as \emph{north} and \emph{west}. As an example, we can have expressions such as \emph{a blue square next to a red circle} or \emph{a blue square north of a red circle}.

To ensure that interpreting the spatial relation is necessary to correctly identify the target object,
we create \emph{visual distractors} in the environment. In the above examples, a blue square that is north of a magenta circle would be a distractor to the intended blue square which is next to the red circle. The existence of visual distractors will force the agent to examine the correspondence between the visual information and the (compositional) language expression to disambiguate and locate the correct target object.

Similar to the gSCAN setup, we use a shared training set and hold-out specific examples to evaluate generalization abilities for learning the visually grounded spatial relation reasoning as in Figure~\ref{fig:split}. In addition to random split (\RN{1}: Random), we create new test splits  including novel object properties (\RN{2}: Visual), novel target and reference combinations (\RN{3}: Relation), novel referent (\RN{4}: Referent), and novel relative positions (\RN{5}, \RN{6}: Relative position).

Table~\ref{tab:relation_split_details} shows detailed descriptions of the held-out examples for each test split.  In total, we generated about 260,000 examples for the new task, using 18 templates and about 32,000 unique instructions.\footnote{The code and data are available at \url{https://github.com/google-research/language/tree/master/language/gscan}.} Details are included in Appendix \ref{sec:data_generation}.

\begin{table}[t]
\centering
\resizebox{\linewidth}{!}
{
    \scriptsize
    \tabcolsep 2pt
    \begin{tabular}{@{}ll@{}}
    \toprule
    Split & Held-out Examples \\
    \midrule
    \RN{1}: Random & Random \\
    \RN{2}: Visual & \emph{Red squares} as target or reference \\
    \RN{3}: Relation & \emph{Green squares} and \emph{blue circles} combinations \\
    \RN{4}: Referent & \emph{Yellow squares} referred as target \\
    \RN{5}: Relative position 1 & Targets are \emph{north} of their references \\
    \RN{6}: Relative position 2 & Targets are \emph{south west} of their references \\
    \bottomrule
    \end{tabular}
}
\caption{Held-out examples for spatial relation learning task.}
\label{tab:relation_split_details}
\end{table}

\paragraph{Results} \llq{Added more discussions from rebuttal} We evaluate the cross-modal attention model on this new task and show results in Table~\ref{tab:exact_match_relation}. The  model outperforms the baseline methods by a large margin on 4 out of 6 splits, but performs surprisingly worse on splits \RN{5} and \RN{6}. The model performs unexpectedly well on the \RN{3}. We conduct ablation studies and report the results in Table~\ref{tab:relation_attention}. \RN{3} is also robust to various ways of removing attentions, similar to the F split we observed in Table~\ref{tab:gscan_attention}. On the other end, the splits \RN{2} and \RN{4} are affected negatively very much, while \RN{5} and \RN{6} are surprisingly improved with removing the cross-modal attention. However, there is not a single model that excels at all splits at the same time.

We hypothesize that the cross-modal attention model overfits to certain aspects of the training distribution, leading to worse out of distribution performance on \RN{5} and \RN{6} splits.
We compute the exact match of our models on examples of seen relations from the random test split, to verify the in-domain generalization. Particularly, this in-domain settings evaluate situations where targets are ``north west/north east” (in-domain \RN{5}) and ``south/west” (in-domain \RN{6}) to their references.
\llq{Added more experiment results from rebuttal}
It shows that the cross-modal attention model (95.52 on in-domain \RN{5} and 95.43 on in-domain \RN{6}) has better in-domain generalization than its without cross-modal counterpart (93.14 on in-domian \RN{5} and 93.90 on in-domain \RN{6}), but it performs worse in compositional generalization. Regularization techniques could potentially be used to improve the cross-modal attention model, which we leave for future research.

\begin{table}
\centering
\resizebox{\linewidth}{!}
{
\small
\tabcolsep 3pt
\begin{tabular}{@{\;}ccccc@{\;}}
\toprule
Split & \makecell{Seq2Seq~\shortcite{ruis2020benchmark}} & \makecell{FiLM~\shortcite{perez2018film}} & \makecell{RN~\shortcite{santoro2017simple}} & \makecell{Ours} \\
\midrule
\RN{1} & 86.48 \scriptsize{$\pm$ 0.64} & 88.85 \scriptsize{$\pm$ 0.82} & 85.17 \scriptsize{$\pm$ 3.81} & \textbf{94.66 \scriptsize{$\pm$ 0.24}} \\
\RN{2}  & 40.10 \scriptsize{$\pm$ 0.83} & 50.68 \scriptsize{$\pm$ 0.32} & 38.59 \scriptsize{$\pm$ 0.74} & \textbf{64.41 \scriptsize{$\pm$ 4.52}} \\
\RN{3} & 86.08 \scriptsize{$\pm$ 0.73} & 88.81 \scriptsize{$\pm$ 1.42} & 85.66 \scriptsize{$\pm$ 4.35} & \textbf{94.89 \scriptsize{$\pm$ 0.20}}\\
\RN{4} & 5.47 \scriptsize{$\pm$ 0.09} & 10.78 \scriptsize{$\pm$ 3.47} & 4.85 \scriptsize{$\pm$ 0.86} & \textbf{49.58 \scriptsize{$\pm$ 3.47}} \\
\RN{5} & \textbf{81.41 \scriptsize{$\pm$ 1.03}} & 76.20 \scriptsize{$\pm$ 2.64} & 79.86 \scriptsize{$\pm$ 3.16} & 59.29 \scriptsize{$\pm$ 5.63} \\
\RN{6} & \textbf{81.84 \scriptsize{$\pm$ 1.38}} & 75.05 \scriptsize{$\pm$ 3.63} & 80.93 \scriptsize{$\pm$ 2.76} & 49.50 \scriptsize{$\pm$ 6.49} \\
\bottomrule
\end{tabular}
}
\caption{Percentage of exact matches by various methods on proposed spatial relation splits.}
\label{tab:exact_match_relation}
\end{table}

\begin{table}
\centering
{
\small
\tabcolsep 2pt
\begin{tabular}{@{\;}ccccc@{\;}}
\toprule
Split &  Ours & $-$V & $-$T & $-$X \\
\midrule
\RN{1} & 94.66 \scriptsize{$\pm$ 0.24} & \textbf{94.70 \scriptsize{$\pm$ 0.23}} & 94.58 \scriptsize{$\pm$ 0.26} & 92.71 \scriptsize{$\pm$ 0.90} \\
\RN{2} & \textbf{64.41 \scriptsize{$\pm$ 4.52}} & 62.42 \scriptsize{$\pm$ 7.12} & 59.75 \scriptsize{$\pm$ 4.18} & 46.65 \scriptsize{$\pm$ 2.20} \\
\RN{3} & 94.89 \scriptsize{$\pm$ 0.20} & 94.90 \scriptsize{$\pm$ 0.25} & \textbf{94.91 \scriptsize{$\pm$ 0.30}} & 92.88 \scriptsize{$\pm$ 0.79} \\
\RN{4} & \textbf{49.58 \scriptsize{$\pm$ 3.47}} & 36.15 \scriptsize{$\pm$ 16.42} & 35.07 \scriptsize{$\pm$ 13.73} & 10.76 \scriptsize{$\pm$ 1.76} \\
\RN{5} & 59.29 \scriptsize{$\pm$ 5.63} & 58.95 \scriptsize{$\pm$ 6.48} & 59.21 \scriptsize{$\pm$ 6.39} & \textbf{86.34 \scriptsize{$\pm$ 1.50}} \\
\RN{6} & 49.50 \scriptsize{$\pm$ 6.49} & 49.40 \scriptsize{$\pm$ 5.66} & 47.91 \scriptsize{$\pm$ 8.28} & \textbf{87.10 \scriptsize{$\pm$ 2.15}} \\
\bottomrule
\end{tabular}
}
\caption{Ablation studies of our model on new relation reasoning task (-V, -T, and -X: removing decoder's visual, decoder's textual, and encoder's cross-modal attentions respectively).}
\label{tab:relation_attention}
\end{table}

\section{Sample Complexity}

\begin{figure*}[t]
    \centering
    \includegraphics[width=\linewidth]{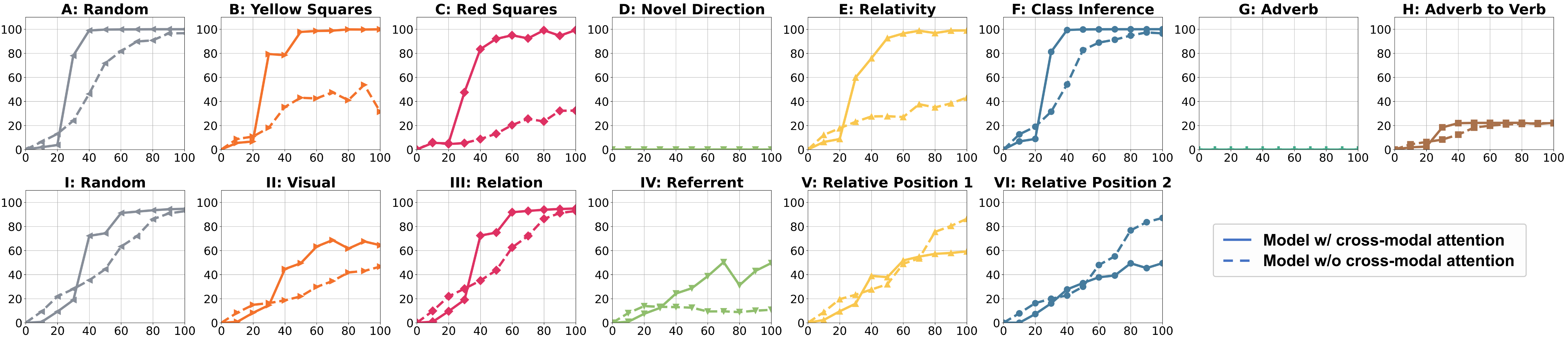}
    \caption{Data efficiency of the models on compositional splits (top) and spatial relation splits (bottom). The x-axis is the percentage of data (\%) and the y-axis is the exact match (\%).}
    \label{fig:data_efficiency}
\end{figure*}

One inspiration of studying systematic generalization is to develop techniques that can reduce the sample complexity for learning novel behaviors. While we observe that many models perform well at least on some splits, we ask a natural question that has not been studied before: \emph{even on those splits, are our models fulfilling the promise of learning systematic generalization?}

\llq{Combined the original with the additional sample complexity experiments in appendix}
Figure ~\ref{fig:data_efficiency} \llq{Updated the figure to show each split separately} shows that for the original gSCAN and the newly proposed task, performance for the cross-modal attention model starts to significantly drop when trained using less than around 40\% of the training data for most splits. The model without cross-modal attention is even more data inefficient; performance starts to significantly drop when trained using less than 70\% of the training data for the original compositional splits, or when reducing the training data by any amount for the spatial relation splits.
This suggests exploring model architectural priors can help improve data efficiency and provide more benefits for generalization.

To investigate how the number of primitives influences the data efficiency, we re-generate smaller compositional splits and spatial relation splits by reducing the number of primitives.
For compositional splits, we exclude one noun (cylinder), one color adjective (blue), and one adverb (while zigzagging).
For spatial relation splits, we exclude one noun (cylinder) and one location preposition (next to).
This reduces the number of training examples by around 2/3 and leads to around 110,000 and 74,000 training examples for the smaller compositional splits and spatial relation splits respectively.

\begin{figure}[ht!]
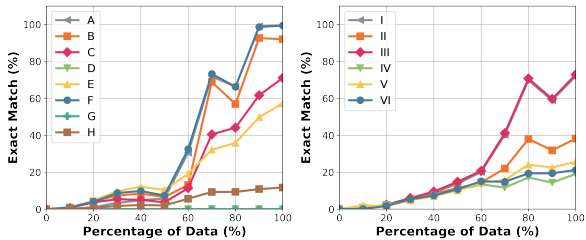

    \centering
    \begin{tabular}{@{}c@{}c@{}}
        \includegraphics[width=0.5\linewidth]{figures/data_compositional_small_x_attn.pdf} &
        \includegraphics[width=0.5\linewidth]{figures/data_relation_small_x_attn.pdf}
    \end{tabular}
    \caption{Data efficiency of the model with cross-modal attention on compositional splits (left) and spatial relation splits (right) with smaller number of primitives.}
    \label{fig:data_efficiency_small}
\end{figure}

We then perform similar experiments with our cross-modal attention model.
The results are shown in Figure \ref{fig:data_efficiency_small}.
When the number of primitives decreases, the percentage of training data required to attain satisfactory performance is increased, with more than 80\% of training data needed for both splits.
One possible explanation is that the total number of training examples is reduced when the number of primitives decreases. Therefore, the model sees fewer combinations during training, thus needs a higher percentage of data to properly learn visual grounding and compositionality.
This provides further evidence that current model need extensive training data to achieve systematic generalization, demonstrating the necessity of evaluating sample complexity for future work.
\section{Related work}

Many datasets and tasks have been proposed for examining systematic generalization: visual question answering~\citep{johnson2017inferring, bahdanau2019closure, pezzelle2019MALeViC, bahdanau2019systematic}, visually grounded navigation instruction following~\citep{hermann2017grounded, yu2018interactive, chevalier2018babyai, chaplot2018gated}.  Many approaches have been proposed too, though a large percentage of them uses task-specific design and only work well for the specific tasks or datasets, for example,  learning neural program executions~\citep{andreas2016neural, hu2017learning, johnson2017inferring, santoro2017simple, hudson2018compositional, mao2019neuro},
in addition to the ones we have described previously.

This work focuses on using a generic cross-modal attention model to probe the gSCAN dataset in the hope of understanding in what aspect the task/dataset is challenging. \citet{ding2020object} also presented new evidences that a neural-based model can solve similar CG tasks. Our interest is to use such models to inspire new task designs that continue to challenge neural models.
\section{Conclusion}

In this work, we have demonstrated that a general-purpose cross-modal attention model can achieve strong performance on a majority of gSCAN splits and outperform more specialized prior work. We have proposed a challenging additional task for gSCAN that requires agents to reason over spatial relations between objects in the visual scene, and have highlighted data efficiency as a consideration for future work.

\section*{Acknowledgements}
We thank Kristina Toutanova and the anonymous reviewers for their helpful feedback and suggestions.

\bibliography{camera_ready}
\bibliographystyle{acl_natbib}

\clearpage
\begin{center}
    \Large \textbf{Appendix}
\end{center}
\appendix
In this appendix, we include more details about  our cross-modal attention model (\S~\ref{sec:model}) and data generation procedure (\S~\ref{sec:data_generation}).

\section{Models}
\label{sec:model}
In this section, we include more details about our cross-modal attention model to ensure all results are fully reproducible. We also include details of the baseline methods we implement.

\subsection{Model details}
\begin{figure}[ht]
\centering
\includegraphics[width=\linewidth]{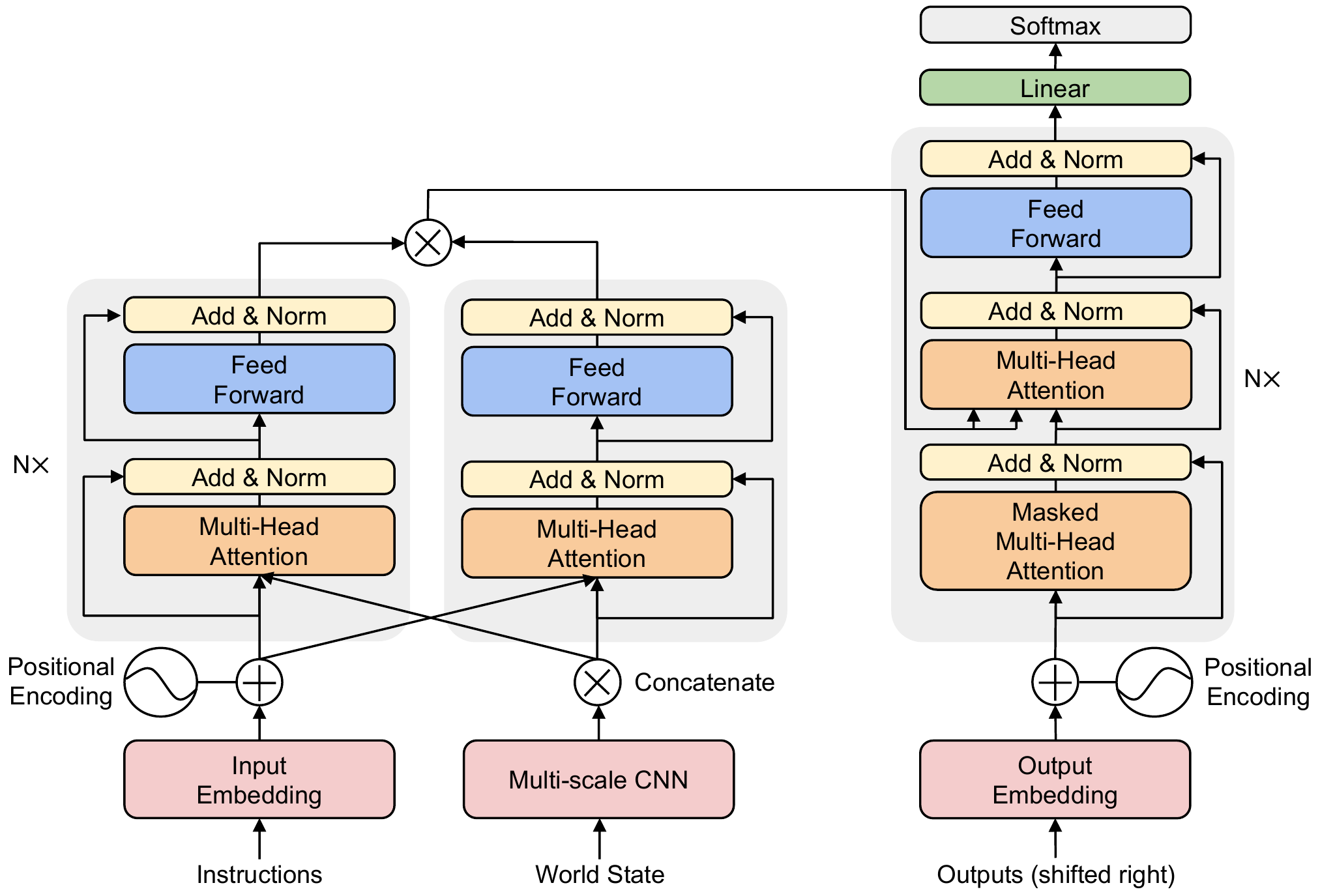}
\caption{The architecture of our model. The encoder consists of 6 layers of transformer block with cross-modal attention. The decoder contains 6 layers of transformer block, and each block has one self-attention block and one multi-head attention block over contextual representations of encoder. \footnotemark}
\label{fig:model}
\end{figure}

We show our model architecture in Figure~\ref{fig:model}.
The model follows the standard encoder-decoder structure in Transformer~\citep{vaswani2017attention} and uses cross-modal attention in the encoder similar to ViLBERT~\citep{lu2019vilbert}.
It is trained via teacher forcing.
During test time, it generates output actions in the auto-regressive manner.

\paragraph{Encoder}
The model takes input command tokens $\vx = \{x_1, x_2, ..., x_n \}$ and world state $\mW \in \Rbb^{d \times d \times c}$, where $d$ is the grid size and $c$ is world state dimension as in the baseline seq2seq model~\citep{ruis2020benchmark}.
The encoder maps the world state to visual representation $\mH^v$ through multi-scale convolutional networks followed by linear layers.
It encodes the input tokens to input embeddings $\mH^l = \{\vh^l_1, \vh^l_2, ..., \vh^l_n \}$ with positional encoding.

The visual and linguistic representations are passed through $N=6$ transformer blocks with cross-modal attention.
Similar to ViLBERT~\citep{lu2019vilbert}, each transformer block consists of two parallel multi-head attention blocks, with the representation of one modality passed as key and value to the multi-head attention block of the other modality.

\paragraph{Decoder}
The decoder consists of $N=6$ stacked blocks similar to the decoder in Transformer~\citep{vaswani2017attention}.
Each block contains one self-attention block and one multi-head attention block over the contextual embeddings $\mH^c = [\mH^v;\mH^l]$ of the encoder.

\subsection{Implementation Details}
We use $N=6$ layers of transformer blocks for the encoder and the decoder.
Each block has hidden size of 128, intermediate size of 256, and 8 attention heads.
The total number of parameters of the model is around 3M for all model variants. 
We train our model for 40 epochs with batch size of 128.
We use the Adam optimizer with weight decay.
The initial learning rate is set to be 0.0016 for model with cross-modal attention and 0.0008 for model without it.
We use linear warmup for the first 10\% of training epoch and decrease the learning rate by 10 in the 20th and 30th epoch.
The dev set accuracy and exact match are used to validate model performance.

\footnotetext{The figure is modified from \url{https://github.com/dair-ai/ml-visuals}.}

\subsection{Baseline Methods}
We implement FiLM~\citep{perez2018film} and RN~\citep{santoro2017simple} as our baselines.
The baselines are built upon the seq2seq model.
FiLM learns functions to predict a set of $\beta$, $\gamma$ conditioned on inputs to modulate neural networks.
For FiLM, we add a linear layer to predict $\beta$, $\gamma$ for each convolution layer.
For RN, we use 3 relation layers with hidden size of 256 followed by 2 fully-connected layers to predict object relations.
The hidden states are then used to initialize the LSTM decoder. 
We refer the readers to the original FiLM and RN paper for model details.
Other hyper-parameters remain the same as the original seq2seq baseline.\footnotemark 

\footnotetext{\url{https://github.com/LauraRuis/multimodal_seq2seq_gSCAN}.}

\section{Data Generation}
\label{sec:data_generation}

We build upon the original gSCAN codebase to generate spatial relation data splits.\footnotemark

\footnotetext{\url{https://github.com/LauraRuis/groundedSCAN}.}

\subsection{Input Commands Generation}
The input commands are generated based on context-free grammar (CFG) in Table~\ref{tab:cfg}.
We add new grammar rules and lexicon rules to support object relations.
Since we focus on evaluating relation reasoning, we exclude templates without any object relation.

\begin{table}[ht]
\centering
\small
\tabcolsep 2.5pt
{
    \begin{tabular}{@{\;}l}
    \toprule
    Grammar Rule \\
    \midrule
    ROOT $\rightarrow$ VP \\
    VP $\rightarrow$ VV\textsubscript{i} \emph{`to'} DP \\
    VP $\rightarrow$ VV\textsubscript{t} DP \\
    DP $\rightarrow$ \emph{`a'} NP \\
    NP $\rightarrow$ JJ NP \\
    NP $\rightarrow$ NP PP \\
    PP $\rightarrow$ LOC DP \\
    NP $\rightarrow$ NN \\
    \toprule
    Lexicon Rule \\
    \midrule
    VV\textsubscript{i} $\rightarrow$ \{walk\} \\
    VV\textsubscript{t} $\rightarrow$ \{push, pull\} \\
    NN $\rightarrow$ \{circle, square, cylinder\} \\
    JJ $\rightarrow$ \{red, green, blue, yellow, big, small\} \\
    LOC $\rightarrow$ \{next to\} \\
    LOC $\rightarrow$ \{east of, north of, west of, south of\} \\
    LOC $\rightarrow$ \{north east of, north west of\} \\
    LOC $\rightarrow$ \{south east of, south west of\} \\
    \bottomrule
    \end{tabular}
}
\caption{The CFG used to generate input commands.}
\label{tab:cfg}
\end{table}

\subsection{World State Generation}
We generate the world state with the following constraints:
1) For each target referent $\mT$, there will be one unique reference object $\mR$ next to it;
2) There will be up to $n$ visual distractors $\mV_1, \mV_2, ..., \mV_n$, which have the same size, shape, and color of the target object $\mT$;
3) Each visual distractor $\mV_i$ will or will not have its own reference object $\mO_i$.

Additionally, to avoid ambiguity, when generating visual distractor $\mV_i$, we ensure:
1) If the input command contains abstract relative position (\ie \emph{next}), $\mV_i$ cannot be placed near $\mR$, and $\mO_i$ must be distinct from $\mR$;
2) If the input command contains specific relative position (\eg \emph{north, west}, etc.), $\mV_i$ can be placed anywhere, and $\mO_i$ can be the same as $\mR$.
However, $\mO_i$ cannot have the same relative position to $\mV_i$ as $\mR$ to $\mT$. Other procedures remain the same as the original setup.

\subsection{Data Examples}
\begin{figure}[H]
    \centering
    \begin{tabular}{@{}c@{\;}c@{}}
        \includegraphics[width=0.5\linewidth]{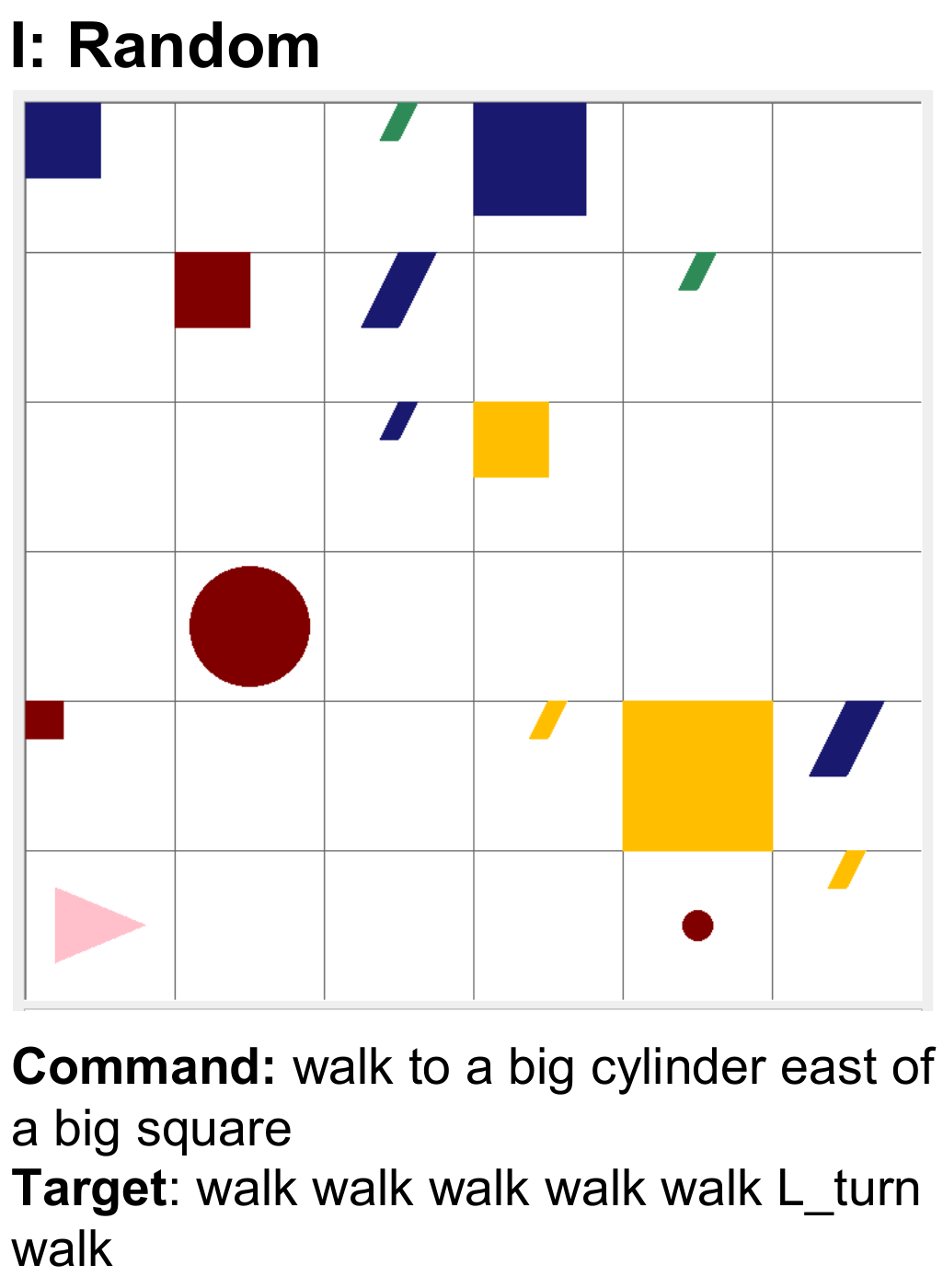} & 
        \includegraphics[width=0.5\linewidth]{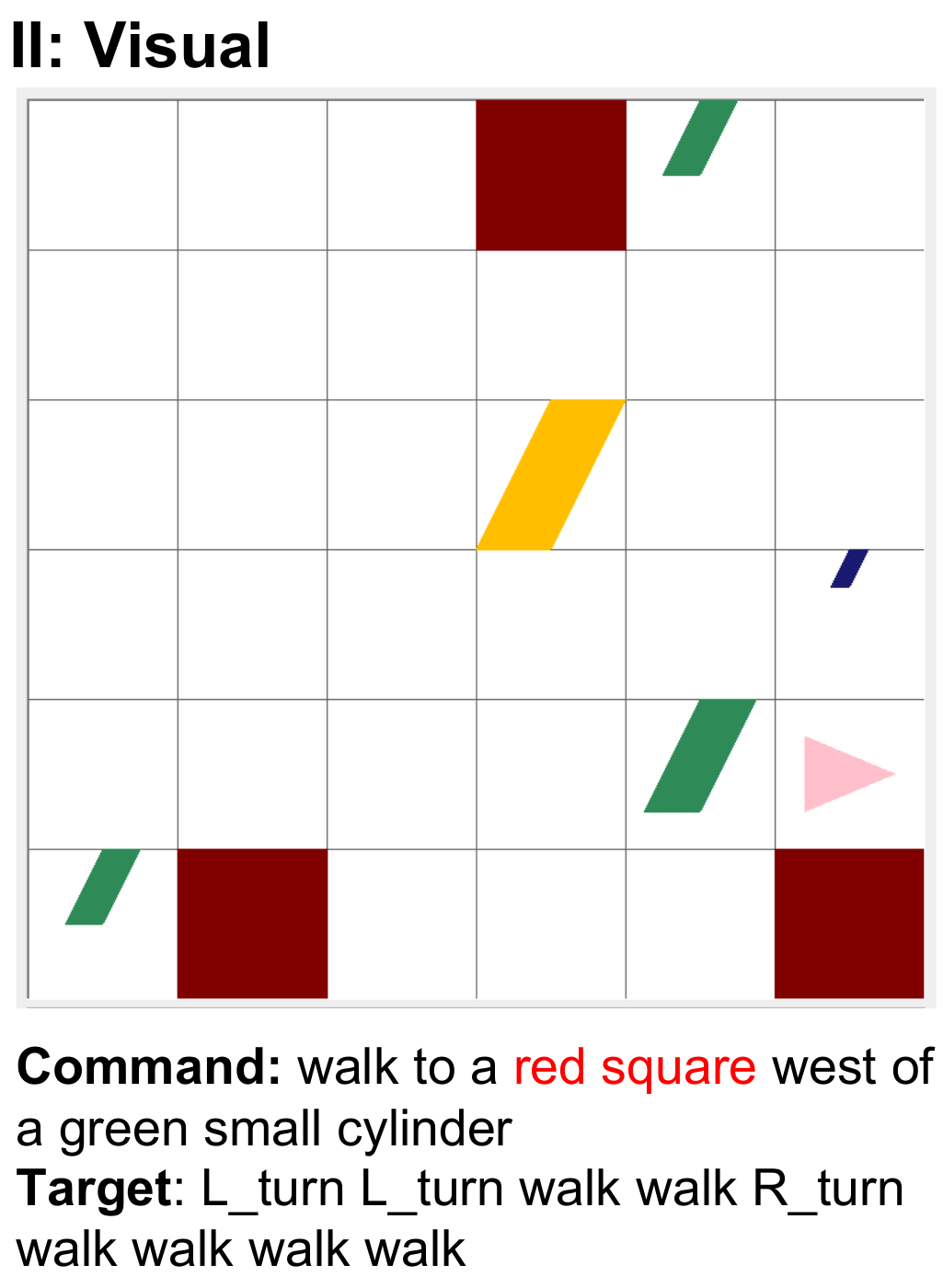} \\
        \includegraphics[width=0.5\linewidth]{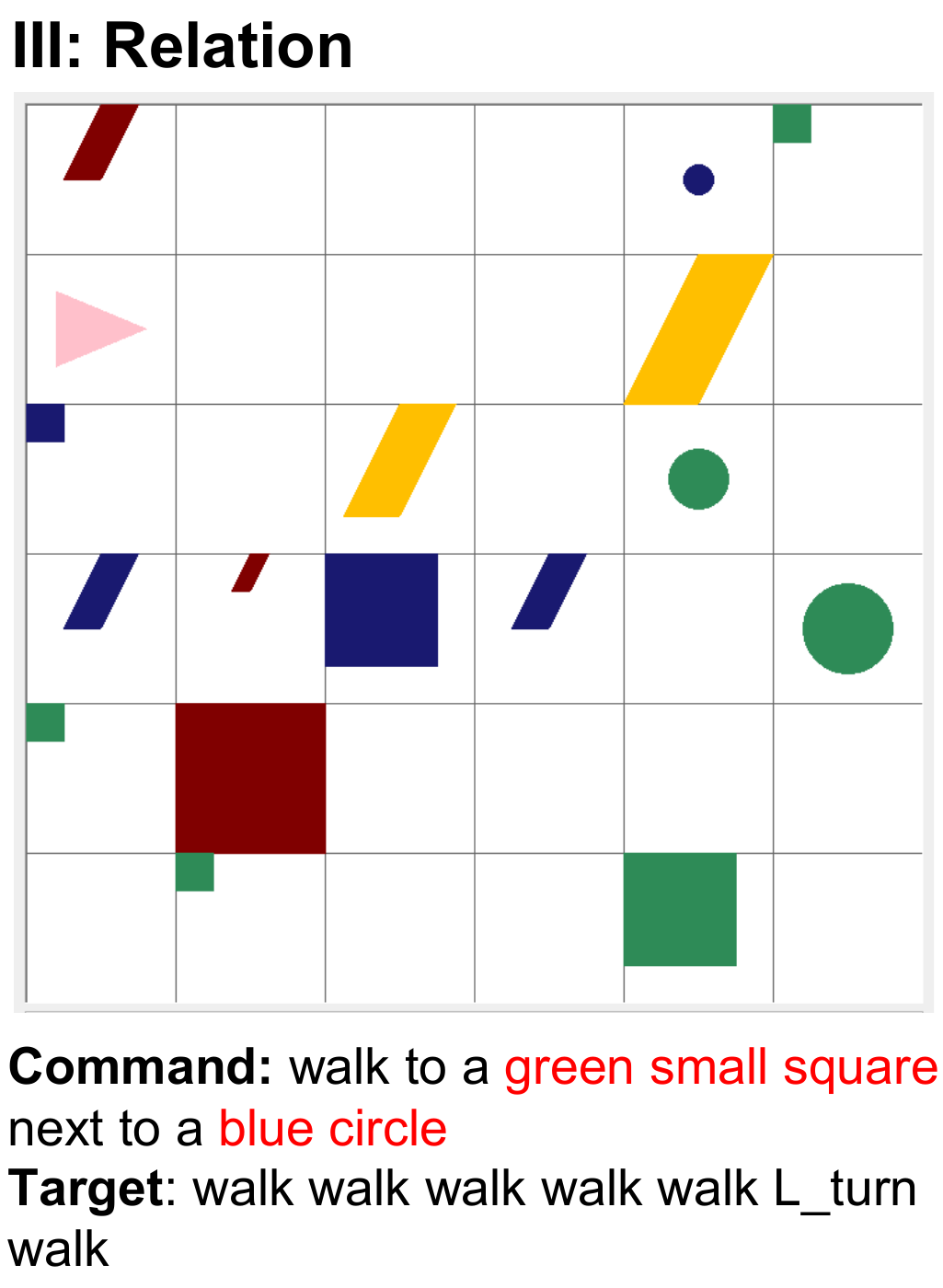} &
        \includegraphics[width=0.5\linewidth]{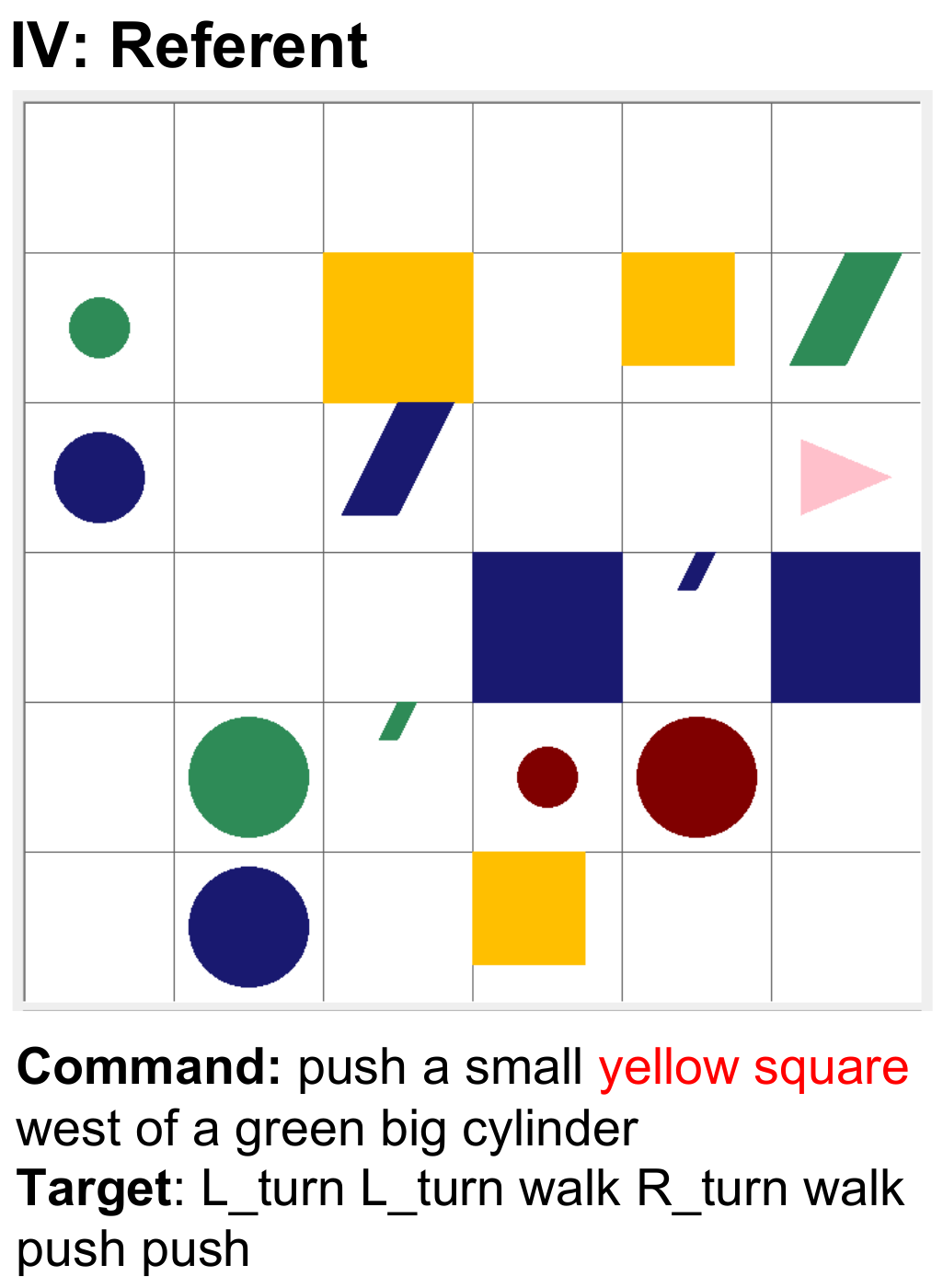} \\ 
        \includegraphics[width=0.5\linewidth]{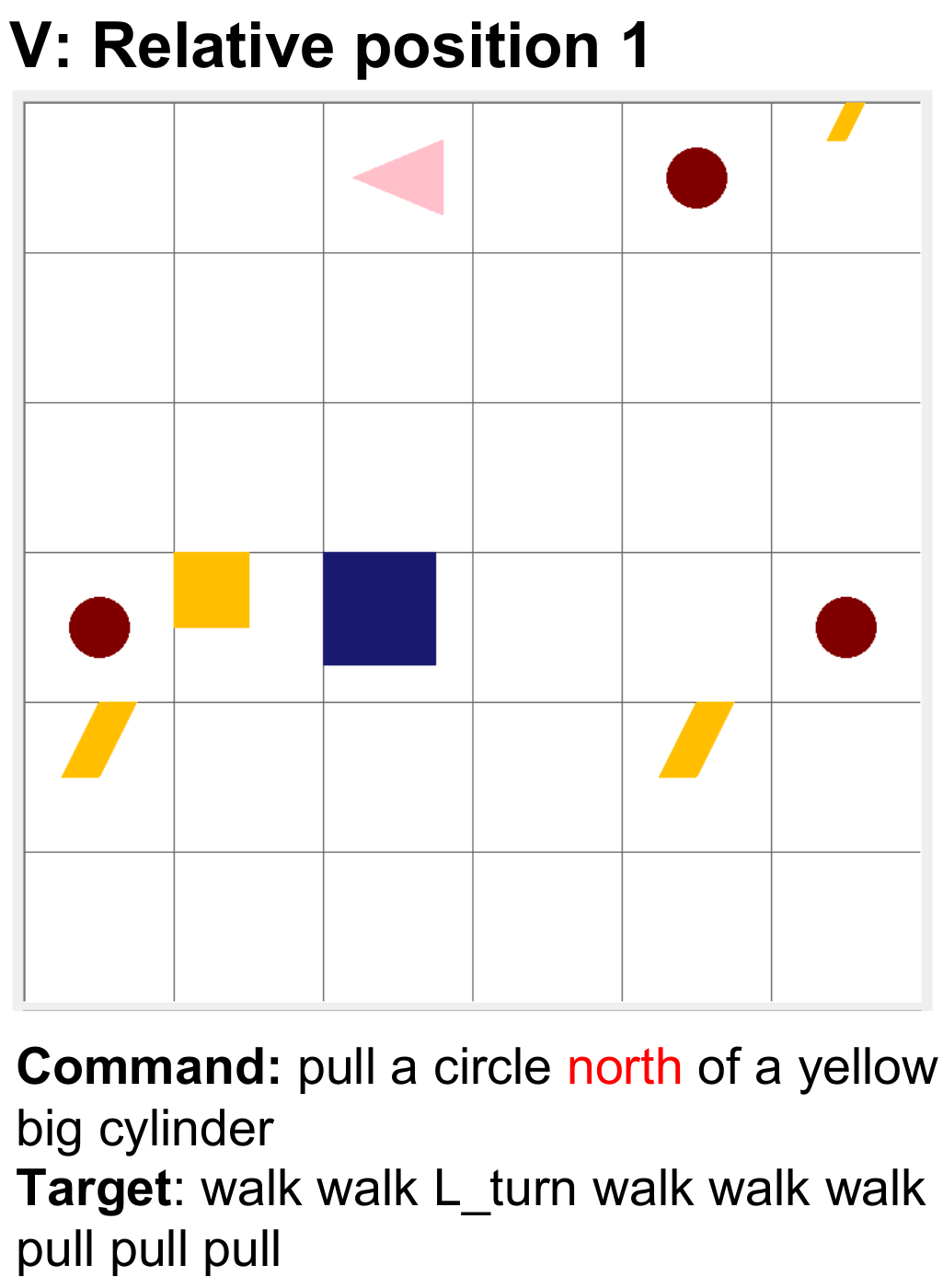} &
        \includegraphics[width=0.5\linewidth]{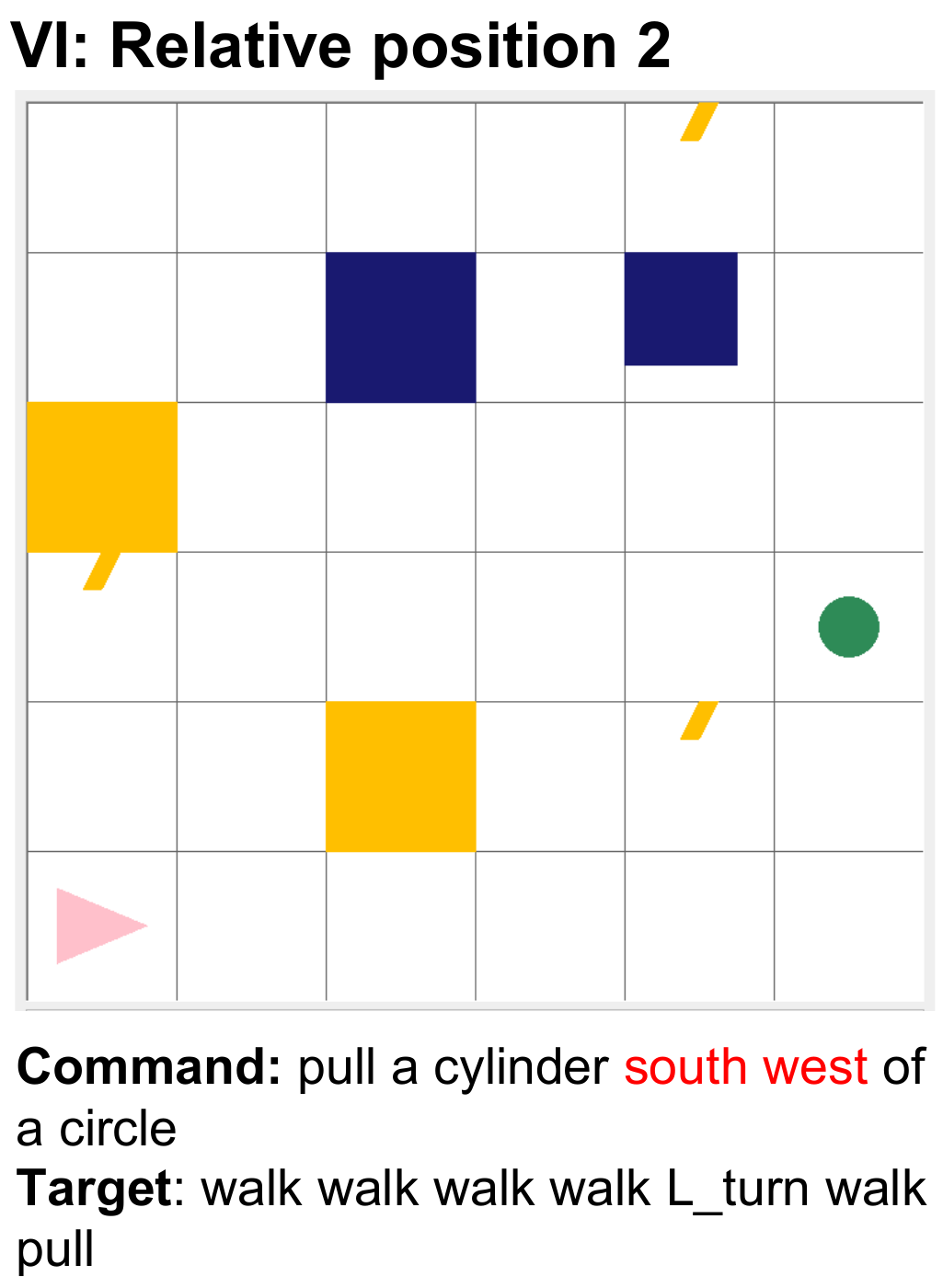}
    \end{tabular}
    \caption{Data examples for each spatial relation test split with the systematic difference highlighted.}
    \label{fig:split_examples}
\end{figure}

We show data examples for each spatial relation split in Figure~\ref{fig:split_examples}.
The systematic difference between the training and test split is highlighted.

\end{document}